\documentclass[a4paper,conference]{IEEEtran}

\ifCLASSINFOpdf

\else

\fi

\usepackage{graphicx}
\usepackage{amsmath}
\usepackage{amssymb}
\usepackage{booktabs,ragged2e}
\usepackage[flushleft]{threeparttable}

\usepackage{subfigure}
\usepackage{tikz}
\usepackage{pifont}
\usepackage{url}
\usepackage{hyperref}
\usepackage{arydshln}

\hypersetup{
    colorlinks=true,
    linkcolor=blue,
    filecolor=magenta,      
    urlcolor=purple,
    pdftitle={Overleaf Example},
    pdfpagemode=FullScreen,
    }

\begin{document}
%
\title{Tuning Vision Foundation Model via Test-Time Prompt-Guided Training for VFSS Segmentations \\}

\author{
    \IEEEauthorblockN{Chengxi Zeng\IEEEauthorrefmark{1},
    David Smithard\IEEEauthorrefmark{3}, 
    Alberto M Gambaruto\IEEEauthorrefmark{1}, 
    Tilo Burghardt\IEEEauthorrefmark{1}
    }
    \IEEEauthorblockA{\IEEEauthorrefmark{1} University of Bristol, UK}
    \IEEEauthorblockA{\IEEEauthorrefmark{3}Queen Elizabeth Hospital, Woolwich, UK\vspace{-20pt}}
    
}

    

\maketitle

\begin{abstract}
Vision foundation models have demonstrated exceptional generalization capabilities in segmentation tasks for both generic and specialized images. However, a performance gap persists between foundation models and task-specific, specialized models. Fine-tuning foundation models on downstream datasets is often necessary to bridge this gap. Unfortunately, obtaining fully annotated ground truth for downstream datasets is both challenging and costly.
To address this limitation, we propose a novel test-time training paradigm that enhances the performance of foundation models on downstream datasets without requiring full annotations. Specifically, our method employs simple point prompts to guide a test-time semi-self-supervised training task. The model learns by resolving the ambiguity of the point prompt through various augmentations. This approach directly tackles challenges in the medical imaging field, where acquiring annotations is both time-intensive and expensive.
We conducted extensive experiments on our new Videofluoroscopy dataset (VFSS-5k) for the instance segmentation task, achieving an average Dice coefficient of 0.868 across 12 anatomies with a single model.
\end{abstract}
%
\vspace{-10pt}\section{Introduction}
Dysphagia, or difficulty swallowing, affects 30–50\% of stroke patients~\cite{Smithard2007LongtermOA} and up to 84\% of older adults with dementia, posing significant risks such as malnutrition, pneumonia, aspiration, and an increased association with mortality~\cite{Smithard2016DysphagiaAG, Smithard1997TheNH}. Early detection and treatment are therefore critical to improving patient outcomes.

The Videofluoroscopic Swallow Study (VFSS) is widely regarded as the gold standard for dysphagia assessment. During VFSS, patients swallow texture-modified foods and liquids containing barium, enabling visualization of bolus trajectory, muscle and hyoid bone movements, and the relationship between anatomical function and aspiration~\cite{Ramsey2003EarlyAO}. Consequently, accurate segmentation of VFSS datasets is of immense clinical value to therapists for diagnosing and managing dysphagia.

Vision foundation models such as Segment Anything (SAM)~\cite{Kirillov2023SegmentA, Ravi2024SAM2S}, have demonstrated state-of-the-art performance in natural image and video segmentation, driven by their ability to generalize across diverse visual domains. Extensions of these models, such as MedSAM~\cite{Ma2023SegmentAI, Zhu2024MedicalS2}, have been introduced to tackle the challenges inherent to medical image segmentation. Despite their broad applicability across medical datasets and imaging modalities (e.g., CT, MRI, histopathology), empirical evidence suggests that these models can underperform when applied to specific modalities compared to specialist models~\cite{Ma2023SegmentAI}. Consequently, domain-specific adaptations of MedSAM through fine-tuning~\cite{Wu2023MedicalSA} or the integration of lightweight, modality-specific adaptors~\cite{Mazurowski2023SegmentAM} have become standard practices. However, such approaches demand extensive, expert-labelled datasets, imposing significant computational and annotation costs.

Another line of research explores Test-Time Training (TTT), also known as Test-Time Adaptation, as a method for training models without requiring labeled data. TTT is a \textit{Transfer Learning} technique that adapts pretrained models from a source domain (SD) to a new target domain (TD) by fine-tuning on a small number of samples. Prominent studies on TTT~\cite{Gandelsman2022TestTimeTW, Liu2021TTTWD, Liu2024DepthAwareTT} have demonstrated that carefully designed self-supervised objective functions enable models to adapt dynamically to distribution shifts, resulting in improved performance on test samples. These methods are particularly well-suited for medical imaging datasets, where test image labels are not immediately available~\cite{Karani2020TestTimeAN}.

Our work aims to identify an effective self-supervised framework for transferring foundation models to specialized domains. Existing research has explored various cues for self-supervised learning, including consistency cues from simple augmentations~\cite{Sun2019TestTimeTW}, denoising cues via denoising autoencoders~\cite{Karani2020TestTimeAN}, reconstruction cues via masked autoencoders~\cite{Gandelsman2022TestTimeTW}, and depth cues~\cite{Liu2024DepthAwareTT}. While many of these methods can be adapted, approaches relying on depth cues are inherently challenging to apply to 2D medical images due to the absence of depth information.
Meanwhile, the unique architecture of vision foundation models, which includes both an image encoder and a prompt encoder, remains underexplored. The prompt encoder was first introduced in SAM~\cite{Kirillov2023SegmentA}, enabling user input prompts such as points or bounding boxes. For single-object segmentation, a point prompt can be ambiguous, as it may correspond to multiple overlapping objects (e.g., organs in medical images). In contrast, a bounding box prompt provides greater precision by restricting segmentation to a specific area and focusing on a single object.
Our work leverages this characteristic of the prompt encoder for self-supervised learning. We hypothesize that applying augmentations to the same image with a single point prompt can produce varied segmentations due to inherent ambiguity. These ambiguity cues can then be incorporated into training through a contrastive loss. This novel paradigm allows users to input simple point prompts during test time, enabling more effective domain adaptation.

Our key contributions are summarised as follows: 
\begin{itemize}
\item{We introduce a novel Test-Time Training paradigm for foundation models that explicitly incorporates prompts into its workflow, enabling semi-self-supervised test-time adaptation through a simple point prompt.}
\item{We introduced a new dataset, VFSS-5k, comprising over 5,000 accurate labels for videofluoroscopy data. The dataset includes annotations for 10 different anatomical structures, such as the pharynx, mandibles, and C1–C7, among others.}
\item{We conducted extensive experiments comparing various self-supervision setups and training strategies, achieving state-of-the-art (SOTA) performance relative to previous TTT methods and narrowing the performance gap to specialist models.}
\end{itemize}

\section{Preliminary and Related Work}
\subsection{Preliminary on Foundation Models}
SAM~\cite{Kirillov2023SegmentA} is a foundation model trained on the extensive SA-1B dataset, which comprises over 11M high-resolution images with an average of 100 detailed mask annotations per image. Its architecture includes a Vision Transformer (ViT)-based encoder, a flexible prompt encoder (e.g., points, bounding boxes, masks), and a decoder for full-resolution segmentation. This large-scale training enables SAM to generalize effectively in zero-shot vision tasks. More recently, SAM2~\cite{Ravi2024SAM2S} expanded its functionality to video object segmentation and identity tracking, leveraging the HieraViT encoder~\cite{Ryali2023HieraAH} and a memory module for instance tracking.
These models have also found applications in medical imaging~\cite{Ma2023SegmentAI, Zhu2024MedicalS2, Ma2024SegmentAI}. Building on SAM, MedSAM~\cite{Ma2023SegmentAI} was specifically designed for medical image segmentation by fine-tuning the model on a large-scale medical image dataset containing 1,570,263 image-mask pairs. This dataset spans 10 imaging modalities and over 30 cancer types, addressing the unique challenges posed by various medical imaging modalities.

\subsection{Preliminary on Test-Time Training (TTT)}
Test-Time Training (TTT)~\cite{Karani2020TestTimeAN, Sun2019TestTimeTW} aims to adapt a pre-trained model to a target distribution at inference time, leveraging an auxiliary self-supervised task without requiring labeled data. Consider a network comprising a shared encoder \(E\), a main task decoder \(D_\text{main}\), and an auxiliary task decoder \(D_\text{aux}\). The objective during pre-training is to minimize the following combined loss:  
\begin{equation}
\min_{E, D_\text{main}, D_\text{aux}} \mathcal{L}_\text{main} + \lambda \mathcal{L}_\text{aux},
\end{equation}
where \(\mathcal{L}_\text{main}\) is the primary task loss, \(\mathcal{L}_\text{aux}\) is the auxiliary loss, and \(\lambda\) is a weighting hyperparameter that balances the two components.
At test time, the model adapts by optimizing only the encoder \(E\) using the self-supervised auxiliary loss:
\begin{equation}
\min_{E} \mathcal{L}_\text{aux}.
\end{equation}
While it is possible to fine-tune the auxiliary decoder \(D_\text{aux}\) alongside the encoder \(E\), empirical evidence suggests that fine-tuning \(E\) alone is sufficient and computationally more efficient.

\subsection{Related Work}
\textbf{VFSS Segmentation.} The key anatomy of interest in the VFSS dataset is the interaction between the pharynx and the bolus. Consequently, a primary task is to segment these structures. Caliskan et al.\cite{Caliskan2020AutomatedBD} trained a Mask R-CNN model on swallowing datasets to detect and segment the bolus, achieving a segmentation accuracy of 0.71 IoU. More recently, Bandini et al.\cite{bandini2021automated} employed Grad-CAM and Active Contour algorithms for weakly supervised bolus localization. Zhang et al.\cite{Zhang2021DeepLearningBased} evaluated four deep learning models for segmenting vallecular bolus residues, with their ensemble approach achieving a segmentation accuracy of 0.72. PECI-Net\cite{Park2024PECINetBS} was specifically designed for bolus segmentation.
Zeng et al. proposed state-of-the-art models, Video-TransUNet and Video-SwinUNet, for pharynx and bolus segmentation, achieving DSC scores of 0.87 and 0.89, respectively, leveraging a spatio-temporal transformer. However, all these models are specialist approaches trained exclusively for pharynx and bolus segmentation and are not generalized to analyze other anatomical structures in the VFSS dataset.

\textbf{Test-Time Training for Segmentation.} Test-Time Training (TTT) addresses the challenge of adapting models to unseen target domains during inference, leveraging self-supervised learning. Early works~\cite{Sun2019TestTimeTW} employed auxiliary tasks like rotation prediction or masked autoencoders for test-time adaptation in image classification, while more recent methods~\cite{Gandelsman2022TestTimeTW, Wang2021TentFT} focused on segmentation by optimizing reconstruction-based objectives or entropy minimization to align feature distributions. In medical image segmentation, contrastive learning and style-transfer methods~\cite{Zhang2024PASSTestTimePT} have been used to mitigate domain shifts. These approaches, however, are often limited by their reliance on static, single-image adaptation and struggle to generalize to temporal tasks like video segmentation, where spatial-temporal dependencies are crucial.

\section{Methodology}

\subsection{Framework Overview}
Our proposed framework leverages prompt-based test-time training (Prompt-TTT) to adapt foundation models for segmentation tasks under domain shifts. Specifically, the framework employs two different types of prompts: \textbf{box prompts} for the main segmentation task and \textbf{point prompts} for the auxiliary task. The use of diverse prompts ensures that the model learns robust representations capable of handling ambiguous inputs, enabling efficient adaptation during test time.

The framework consists of:
\begin{itemize}
    \item \textbf{Shared Encoder (\(E\))}: A pre-trained encoder that extracts features from input images.
    \item \textbf{Segmentation Decoder (\(D_\text{seg}\))}: Generates segmentation masks conditioned on box prompts.
    \item \textbf{Auxiliary Decoder (\(D_\text{aux}\))}: Performs point-prompt-based segmentation, serving as the self-supervised auxiliary task.
\end{itemize}

\begin{figure*}[h]
\centering
\hspace{15pt} 
\includegraphics[width=250pt]{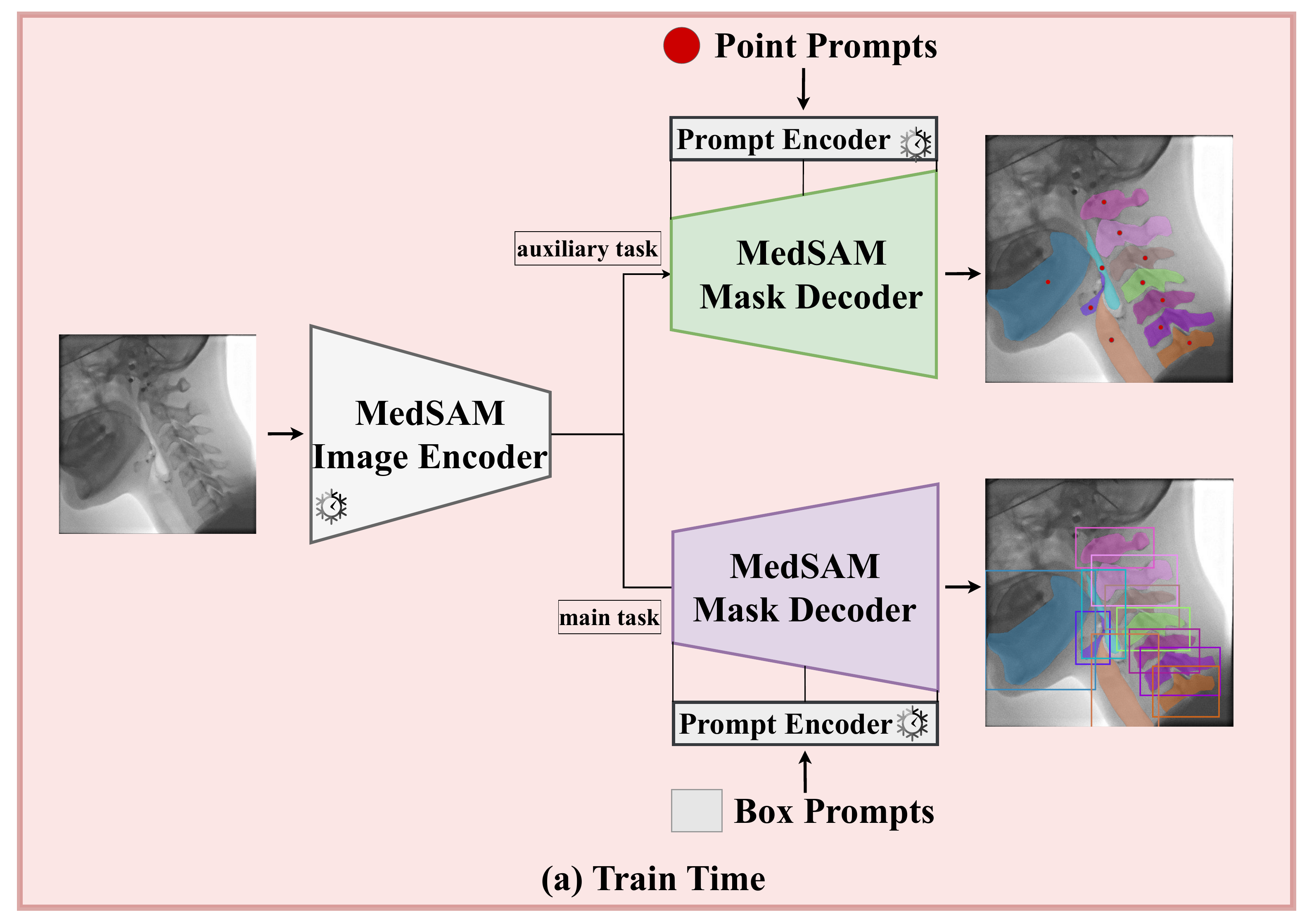}
\hfill
\hspace{-15pt} 
\includegraphics[width=250pt]{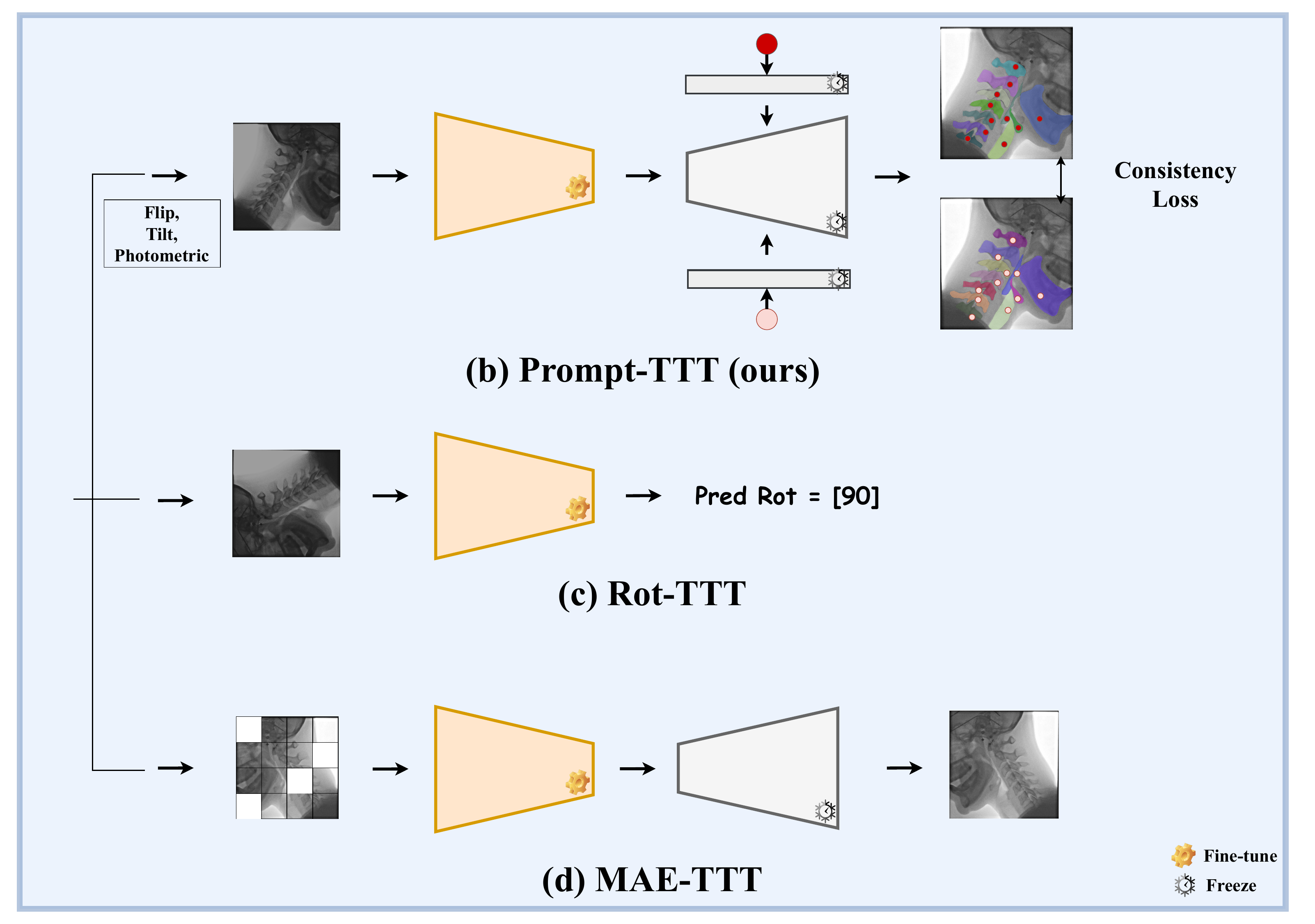}
\caption{Schematic diagram. (a) Training time workflow – the image is processed in two tasks: the main task (box-prompted) and the auxiliary task (point-prompted). (b) Our TTT strategy leverages the resolution of ambiguity generated from different point prompts, in comparison to other TTT methods such as (c) Rotation Prediction and (d) Masked Image Reconstruction.}
\label{fig:reversed_combined_representation}
\end{figure*}

\subsection{Training Phase}
During the training phase, the model is optimized on a labeled source domain dataset with two distinct tasks:
\begin{enumerate}
    \item \textbf{Main Task (Box-Prompt Segmentation)}: 
    The segmentation decoder \(D_\text{seg}\) predicts a mask \(\hat{m}_b\) conditioned on a box prompt \(p_b\). The prediction is given by:
    \[
    \hat{m}_b = D_\text{seg}(E(x), p_b),
    \]
    where \(x\) is the input image, the loss for this task combines Dice Loss and Binary Cross-Entropy (BCE) Loss, both of which are commonly used in standard segmentation tasks:
    \[
    \mathcal{L}_\text{main} = \text{DiceLoss}(m, \hat{m}_b) + \text{BCE}(m, \hat{m}_b),
    \]
    where \(m\) is the ground truth segmentation mask for the box prompt.

    \item \textbf{Auxiliary Task (Point-Prompt Segmentation)}:
    The auxiliary decoder \(D_\text{aux}\) predicts a mask \(\hat{m}_p\) conditioned on a point prompt \(p_p\). The prediction is given by:
    \[
    \hat{m}_p = D_\text{aux}(E(x), p_p).
    \]
    The loss for this task is:
    \[
    \mathcal{L}_\text{aux} = \text{BCE}(m, \hat{m}_p),
    \]
    where \(m\) is the ground truth mask for the point prompt, and \text{BCE} is the binary cross-entropy loss.
\end{enumerate}

The total training loss combines these objectives:
\[
\mathcal{L}_\text{train} = \mathcal{L}_\text{main} + \lambda \mathcal{L}_\text{aux},
\]
where \(\lambda\) is a hyperparameter that balances the contributions of the two losses. Empirically, $\lambda$ is chosen as 0.2 in our experiments.

\subsection{Test-Time Training}
At test time, the framework adapts to unseen target domain data by ensuring consistent predictions for the same image under different transformations and point prompts. This consistency is enforced by fine-tuning the shared encoder \(E\) while keeping the decoders frozen.

An input image \(x\) is augmented with different transformations, such as random rotations, flips, or intensity changes, resulting in transformed versions \(x_1\) and \(x_2\).
Both transformed images are passed through the shared encoder \(E\), along with different point prompts \(p_{p1}\) and \(p_{p2}\):
    \[
    \hat{m}_1 = D_\text{aux}(E(x_1), p_{p1}),
    \]
    \[
    \hat{m}_2 = D_\text{aux}(E(x_2), p_{p2}),
    \]
    where \(D_\text{aux}\) is the auxiliary decoder that predicts segmentation masks based on point prompts.
The predictions \(\hat{m}_1\) and \(\hat{m}_2\) are constrained to be consistent using a consistency loss:
    \[
    \mathcal{L}_\text{consistency} = \| \hat{m}_1 - \hat{m}_2 \|_2^2,
    \]
    where \(\| \cdot \|_2^2\) represents the mean squared error between the two predicted masks.
The encoder \(E\) is fine-tuned by minimizing the total test-time loss.

\subsection{Test-Time Training Through the Video}
Unlike other medical images, VFSS consists of multiple temporal frames during the swallowing process. For a video sequence \(V = \{v_1, v_2, \dots, v_T\}\), the encoder \(E\) is updated iteratively over \(K\) loops:
\[
E^{(k)} = \arg \min_{E} \sum_{t=1}^T \mathcal{L}_\text{total}^t,
\]
with weights from the previous loop \(k-1\) initializing the next loop.

\begin{table*}[h]
\caption{\textbf{Quantitative Results.} Pharynx Segmentation results comparing specialist models to foundation models.} 
\vspace{-10pt}
\label{tab:comparison method}
\begin{center}       
\begin{scriptsize}    
\begin{tabular}{l|c|c|c|c|c|c} 
\toprule
\rule[-1ex]{0pt}{3.5ex}  \textbf{Models (Year)} & \textbf{FLOPs} & \textbf{\#Params} & \textbf{DSC}$\uparrow$ & \textbf{HD95}$\downarrow$ & \textbf{ASD}$\downarrow$ & \textbf{Sensitivity}$\uparrow$\\
\midrule 
\rule[-1ex]{0pt}{3.5ex}  UNet\cite{Ronneberger2015UNetCN} (2015) & 50.1G & 34.5M & $0.842$ & $14.753$ & $2.168$ & $0.829$\\
\rule[-1ex]{0pt}{3.5ex}  NestedUNet\cite{Stoyanov2018DeepLI} (2018) & 105.7G & 36.6M & $0.834$ & $13.760$ & $2.228$ & $0.831$\\
\rule[-1ex]{0pt}{3.5ex}  ResUNet\cite{Zhang2017RoadEB} (2017) & 43.1G & 31.5M & $0.847$ & $11.982$ & $2.049$ & $0.818$\\
\rule[-1ex]{0pt}{3.5ex}  AttUNet\cite{Oktay2018AttentionUL} (2018) & 51.0G & 34.8M & $0.850$ & $12.936$ & $2.183$ & $0.833$\\
\rule[-1ex]{0pt}{3.5ex}  TransUNet\cite{Chen2021TransUNetTM} (2021) & 29.3G & 105.3M & $0.859$ & $7.451$ & $1.105$ & $0.849$\\
\rule[-1ex]{0pt}{3.5ex}  Video-TransUNet\cite{Zeng2022VideoTransUNetTB} (2022) & 40.4G & 110.5M & $0.880$ & $6.916$ & \colorbox{cyan!25}{$1.038$} & $0.885$\\
\rule[-1ex]{0pt}{3.5ex}  SwinUNet\cite{Cao2021SwinUnetUP} (2023) & 6.1G & 27.1M & $0.848$ & $10.290$ & $2.082$ & $0.846$\\
\rule[-1ex]{0pt}{3.5ex}  Video-SwinUNet\cite{zeng2023video} (2023) & 25.8G & 48.9M & \colorbox{cyan!25}{$0.899$} & \colorbox{cyan!25}{$6.237$} & $1.308$ & \colorbox{cyan!25}{$0.901$}\\
\midrule 
\midrule 
\rule[-1ex]{0pt}{3.5ex}  MedSAM\cite{Ma2023SegmentAI} (2023) & 17.7G & 93.7M & $0.813$ & $12.421$ & $2.423$ & $0.820$\\
\rule[-1ex]{0pt}{3.5ex}  MedSAM (Fine-tune only) & 17.7G & 93.7M & $0.841$ & $10.323$ & $2.117$ & $0.833$\\
\rule[-1ex]{0pt}{3.5ex}  MedSAM (Rot-TTT) & 17.7G & 93.7M & $0.852$ & $9.109$ & $1.965$ & $0.851$\\
\rule[-1ex]{0pt}{3.5ex}  MedSAM (MAE-TTT) & 17.7G & 93.7M & $0.864$ & $9.264$ & $1.884$ & $0.860$\\
\rule[-1ex]{0pt}{3.5ex}  MedSAM (Prompt-TTT) (Ours) & 17.7G & 93.7M & \colorbox{blue!25}{$0.881$} & \colorbox{blue!25}{$8.231$} & \colorbox{blue!25}{$1.532$} & \colorbox{blue!25}{$0.885$} \\
\bottomrule
\end{tabular}
\end{scriptsize}    
\end{center}
\vspace{-20pt}
\end{table*}

After \(K\) loops, the updated encoder \(E^{(K)}\) is used for inference. The segmentation decoder \(D_\text{seg}\) remains frozen, and the final mask for frame \(v_t\) is:
\[
\hat{m}_b^t = D_\text{seg}(E^{(K)}(v_t), p_b^t),
\]
where \(p_b^t\) is the box prompt.

This method leverages temporal information and global video context, improving robustness and reducing segmentation inconsistencies across frames.
It has been found to be a more effective TTT strategy than momentum update ones~\cite{Liu2024DepthAwareTT}.

\section{EXPERIMENTS AND RESULTS}

\begin{figure*}
\begin{center}
\begin{tabular}{c} 
\includegraphics[height=6cm]{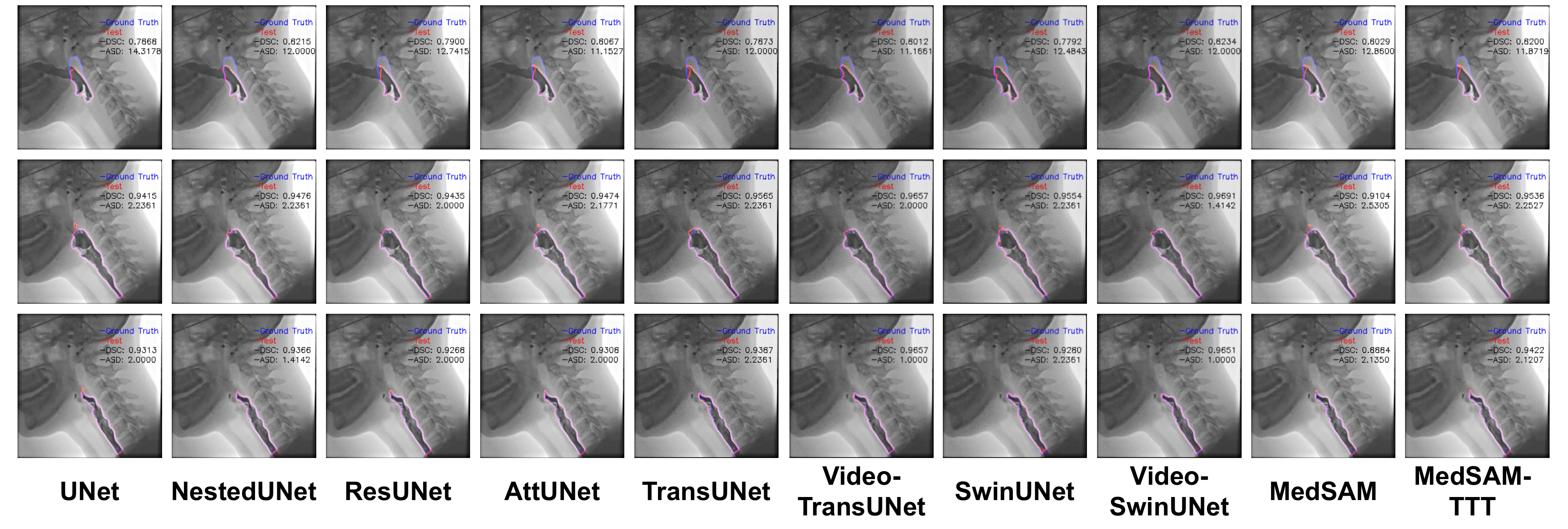}
\end{tabular}
\end{center}
\caption[example]
{\label{fig:qualitative} 
\textbf{Quantitative Results.} Comparing specialist models with the foundation model MedSAM and with our TTT strategies in three consecutive frames.}
\vspace{-00pt}
\end{figure*} 

\subsection{Datasets and Implementation details}
\textbf{Datasets.} The VFSS2022 dataset~\cite{Zeng2022VideoTransUNetTB} is a high-quality swallowing video dataset where patients underwent modified barium swallow tests (VFS studies) under practitioner supervision. VFSS2022 Part 1 comprises 3.5 minutes of swallowing videos sampled at 440 frames (512×512 pixels resolution). Three experts annotated each frame with labels for bolus and pharynx, which were then reviewed by two speech and language therapists. The final ground truth was generated by fusing these annotations using the STAPLE algorithm~\cite{Warfield2004SimultaneousTA}. VFSS5K extends VFSS2022 Part 1 with comprehensive annotations of approximately 12 anatomical structures per frame, including bolus, pharynx, trachea, epiglottis, mandible and cervical vertebrae (C1-C7), totaling around 5,000 instance annotations.\\
\textbf{Implementation details.} Each instance annotations are number coded from 1 - 12 and 0 indicates background. All experiments supported online data augmentation such as random cropping and flipping. The dataset is splitted to 8:2 training and test set ratio. Since we don't use the full annotation during test time training, we only choose random points from the annotations for the point prompt encoder. The foundation model used is the medical segment anything model (MedSAM)~\cite{Ma2023SegmentAI}.  During training (fine-tuning), our system takes in a batch size of 2 and is equipped with an Adam optimizer with an initial learning rate of 1e-3. For test time training, the learning rate is dropped to 6e-5 at the beginning. A learning rate scheduler is set to drop the learning rate to 80\% after 20 epochs of loss saturation. The architecture is achieved in Python 3.9 and Pytorch 2.0 and trained with an NVIDIA Geforce 3090 24GB and a P100 16GB GPU.

\subsection{Comparison with Specialist Models}

We evaluate our proposed architecture against prominent medical image segmentation models, such as UNet\cite{Ronneberger2015UNetCN}, NestedUNet\cite{Stoyanov2018DeepLI}, ResUNet\cite{Zhang2017RoadEB}, AttUNet\cite{Oktay2018AttentionUL}, TransUNet\cite{Chen2021TransUNetTM}, Video-TransUNet\cite{Zeng2022VideoTransUNetTB}, SwinUNet\cite{Cao2021SwinUnetUP}, and Video-SwinUNet\cite{zeng2023video}. These models are compared across four standard evaluation metrics: the Dice Coefficient (DSC), the 95th percentile of the Hausdorff Distance (HD95), the Average Surface Distance (ASD), and Sensitivity, as detailed in Table~\ref{tab:comparison method}. Furthermore, we assess each model's size and computational performance by including the total number of parameters and the total floating-point operations (FLOPs).
It is evident that the unaltered foundation model, MedSAM, exhibits very limited performance on this specific dataset. With fine-tuning on the training set, its performance has improved by a notable margin (2.8\% in DICE). However, it still falls significantly short of the state-of-the-art specialist models (0.899). We explored several test-time training strategies, including Rotation TTT~\cite{Sun2019TestTimeTW} and MAE TTT~\cite{Gandelsman2022TestTimeTW}. The results are presented in Table~\ref{tab:comparison method}.
The two test-time training strategies achieved slight performance improvements but remained limited in effectiveness as they did not fully leverage the prompt encoder. Our approach, however, integrates the prompt into the loop, providing better training guidance during test time.

\subsection{Test-Time performance on different Anatomies}
\begin{table}[h] 
    \caption{\textbf{Quantitative Results.} Foundation model Segmentation performance on different Anatomies, DICE is repoted}  
    \vspace{-10pt} 
    \label{tab:anatomies} 
    \begin{center}        
    \begin{scriptsize}     
    \begin{tabular}{l|c|c|c}  
    \toprule 
    \rule[-1ex]{0pt}{3.5ex}  \textbf{Anatomies} & \textbf{MedSAM} & \textbf{MedSAM (Fine-Tune)} & \textbf{MedSAM (TTT)}\\ 
    \midrule  
   \rule[-1ex]{0pt}{3.5ex}  Bolus & $0.819$ & $0.831$ & \colorbox{blue!25}{$0.865$}\\ 
    \rule[-1ex]{0pt}{3.5ex}  Pharynx & $0.813$ & $0.842$ & \colorbox{blue!25}{$0.881$}\\ 
    \rule[-1ex]{0pt}{3.5ex}  Trachea & $0.837$ & $0.859$ & \colorbox{blue!25}{$0.864$}\\ 
    \rule[-1ex]{0pt}{3.5ex}  Epiglottis & $0.816$ & $0.842$ & \colorbox{blue!25}{$0.852$}\\ 
    \rule[-1ex]{0pt}{3.5ex}  Mandible & $0.852$ & $0.869$ & \colorbox{blue!25}{$0.873$}\\ 
    \rule[-1ex]{0pt}{3.5ex}  Cervical Spine (C1) & $0.826$ & $0.842$ & \colorbox{blue!25}{$0.863$}\\ 
    \rule[-1ex]{0pt}{3.5ex}  Cervical Spine (C2) & $0.845$ & $0.851$ & \colorbox{blue!25}{$0.861$}\\ 
    \rule[-1ex]{0pt}{3.5ex}  Cervical Spine (C3) & $0.846$ & $0.861$ & \colorbox{blue!25}{$0.878$} \\ 
    \rule[-1ex]{0pt}{3.5ex}  Cervical Spine (C4) & $0.833$ & $0.844$ & \colorbox{blue!25}{$0.879$}\\ 
    \rule[-1ex]{0pt}{3.5ex}  Cervical Spine (C5) & $0.824$ & $0.837$ & \colorbox{blue!25}{$0.860$}\\ 
    \rule[-1ex]{0pt}{3.5ex}  Cervical Spine (C6) & $0.830$ & $0.871$ & \colorbox{blue!25}{$0.889$} \\ 
    \rule[-1ex]{0pt}{3.5ex}  Cervical Spine (C7) & $0.817$ & $0.844$ & \colorbox{blue!25}{$0.866$}\\ 
    \midrule  
    \rule[-1ex]{0pt}{3.5ex}  \textbf{Average} & $0.831$ & $0.847$ & \colorbox{blue!25}{$0.868$}\\
    \bottomrule
    \end{tabular} 
    \end{scriptsize}     
    \end{center} 
    \vspace{-10pt} 
    \end{table}
    Table~\ref{tab:anatomies} demonstrates a clear trend of improved segmentation performance across various anatomical structures when transitioning from the baseline MedSAM model to its fine-tuned and test-time training (TTT) variants. The results highlight the benefits of adapting the foundation model to specific datasets and leveraging test-time strategies for further refinement, without the need to extensively train a specialized model for each anatomy.

    The baseline MedSAM model achieves reasonable performance, as seen in its DICE scores, but TTT introduces notable improvements for all anatomies. For instance, the Pharynx's score increases from 0.813 to 0.842, while TTT further elevates it to 0.881, reflecting a 6.8\% improvement.
    
    Overall, the results validate the effectiveness of TTT in enhancing segmentation performance. These strategies are impactful for anatomies with higher variability, paving the way for their application in diverse clinical scenarios.

\subsection{Number of point prompts in TTT}

In this section, we conduct an ablation study on using different point prompts during test-time training. More point prompts lead to less ambiguity but can also limit generalization.
The results in Table~\ref{tab:prompts} show that using a single point prompt achieves the best performance. Increasing the number of prompts to three or five causes a steady decline in performance.
This trend suggests that more prompts during test-time training resolve more ambiguity from the human in the loop. However, this is adverse to the consistency loss computation and results in the model being less effectively trained. This further solidifies our hypothesis that resolving the ambiguity between the two-stream prompt encoder is a good auxiliary task in the test-time training process.

\begin{table}[h]
\caption{\textbf{Quantitative Results.} Ablation study on number of point prompts during Test-Time trainig} 
\vspace{-10pt}
\label{tab:prompts}
\begin{center}       
\begin{scriptsize}    
\begin{tabular}{c|c|c|c|c} 
\toprule
\rule[-1ex]{0pt}{3.5ex}  \textbf{\# of Point Prompts} & \textbf{DSC}$\uparrow$ & \textbf{HD95}$\downarrow$ & \textbf{ASD}$\downarrow$ & \textbf{Sensitivity}$\uparrow$\\
\toprule
\rule[-1ex]{0pt}{3.5ex}  1 & \colorbox{blue!25}{$0.881$} & \colorbox{blue!25}{$8.231$} & \colorbox{blue!25}{$1.532$} & \colorbox{blue!25}{$0.885$}\\
\rule[-1ex]{0pt}{3.5ex}  3 & $0.875$ & $9.332$ & $1.784$ & $0.877$\\
\rule[-1ex]{0pt}{3.5ex}  5 & $0.867$ & $9.994$ & $1.987$ & $0.874$\\
\bottomrule
\end{tabular}
\end{scriptsize}    
\end{center}
\vspace{-10pt}
\end{table}

\section{Conclusion, Limitation and Future work}
We presented an end-to-end framework that successfully exploits multi-frame inputs to segment VFSS data, significantly closing the performance gap with specialist models while also reducing model size. Our proposed neural network effectively utilizes both local and global spatial context and leverages temporal features. Each module within the framework can be fine-tuned or replaced. The final design achieves superior performance compared to other approaches, offering a novel and alternative pipeline for medical video segmentation tasks.
A key limitation of test-time training (TTT) is its challenge in large-scale deployment. Unlike other methods, TTT requires human-provided point prompts for each test sample, which introduces an additional step in the loop. In contrast, other approaches do not have this requirement.
Future work can explore alternative directions in test-time training, such as using specialized models as meta-models to supervise test-time training in foundation models.

\ \\

\newpage

\bibliographystyle{IEEEtran}
\bibliography{IEEEexample}

\end{document}